\pdfoutput=1

\documentclass[11pt]{article}

\usepackage[preprint]{acl}

\usepackage{times}
\usepackage{latexsym}

\usepackage{enumitem}

\usepackage[T1]{fontenc}

\usepackage[utf8]{inputenc}

\usepackage{microtype}

\usepackage{inconsolata}

\usepackage{graphicx}

\usepackage{multirow}
\usepackage{booktabs}
\usepackage{diagbox}
\usepackage{nicematrix}
\usepackage{xcolor,colortbl}
\usepackage{tabularx}
\usepackage{listings}

\definecolor{skyblue}{rgb}{0.53, 0.81, 0.92}
\definecolor{mygreen}{HTML}{EFFEEC}
\definecolor{textualcolor}{HTML}{DAE8FC}
\definecolor{visualcolor}{HTML}{D5E8D4}
\definecolor{crossmodalcolor}{HTML}{F6BDC9}

\newcommand{\q}[1]{``#1''}

\newcommand{\myemoji}{\raisebox{-0.2\height}{\includegraphics[height=1.5em]{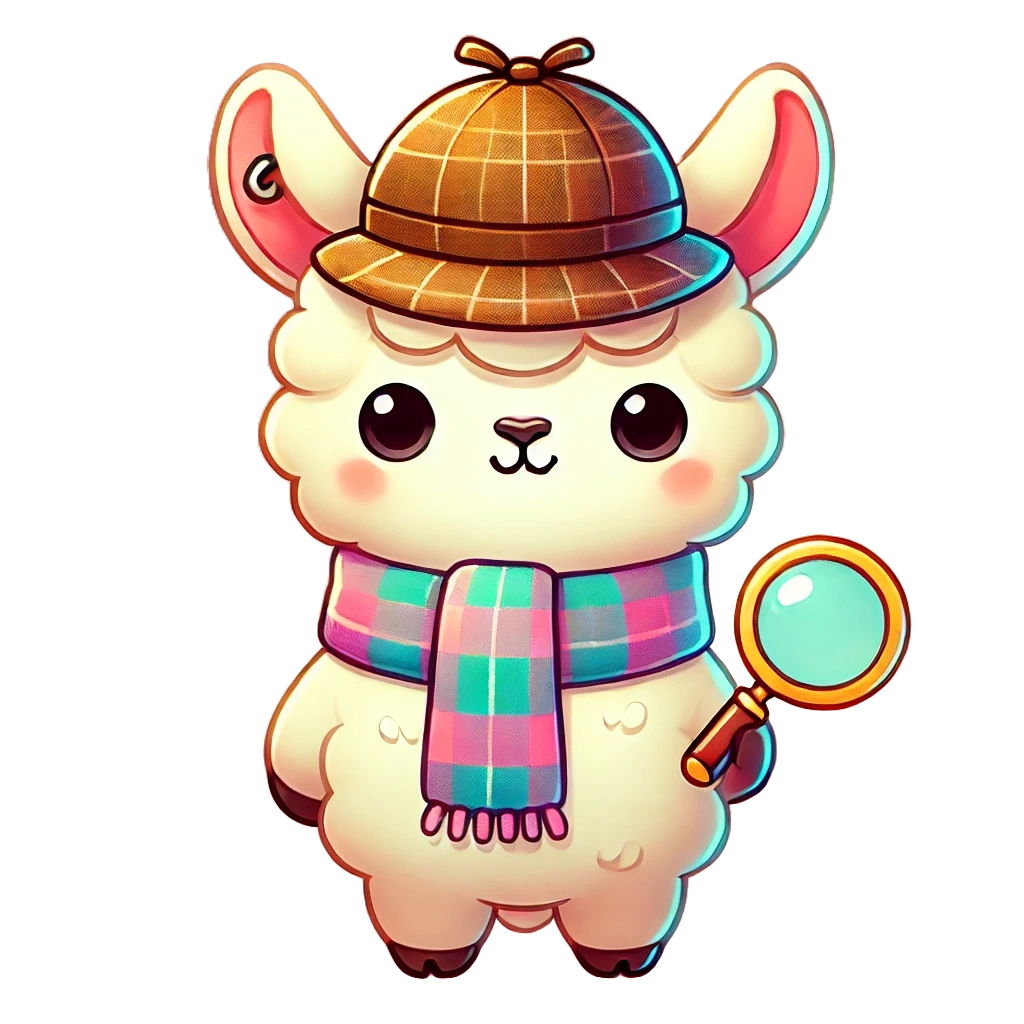}}}

\newcolumntype{C}[1]{>{\centering\arraybackslash}p{#1}}

\usepackage{listings}
\definecolor{darkred}{rgb}{0.6,0.0,0.0}
\definecolor{darkgreen}{rgb}{0,0.50,0}
\definecolor{lightblue}{rgb}{0.0,0.42,0.91}
\definecolor{orange}{rgb}{0.99,0.48,0.13}
\definecolor{grass}{rgb}{0.18,0.80,0.18}
\definecolor{pink}{rgb}{0.97,0.15,0.45}
\definecolor{codegreen}{rgb}{0,0.6,0}
\definecolor{codegray}{rgb}{0.5,0.5,0.5}
\definecolor{codepurple}{rgb}{0.58,0,0.82}
\definecolor{backcolour}{rgb}{0.95,0.95,0.92}
\lstdefinestyle{mystyle}{
  frame=single,
  basicstyle=\ttfamily\footnotesize,
  backgroundcolor=\color{backcolour}, commentstyle=\color{codegreen},
  commentstyle=\color{darkgreen}\slshape,
  keywordstyle=\color{blue},
  stringstyle=\color{darkred},
  numberstyle=\tiny\color{codegray},
  emphstyle=\color{pink}\underbar,
  morekeywords={Verify, Question},
  escapeinside={(*@}{@*)},
  breakatwhitespace=false,         
  breaklines=true,                 
  captionpos=b,                    
  keepspaces=true,                    
  numbersep=5pt,                  
  showspaces=false,                
  showstringspaces=false,
  showtabs=false,                  
  tabsize=2
}

\title{\myemoji{}\hspace{-0.3em}
TRUST-VL: An Explainable News Assistant \\ for General Multimodal Misinformation Detection}

\author{
 \textbf{Zehong Yan},
 \textbf{Peng Qi\thanks{Corresponding author}},
 \textbf{Wynne Hsu} and
 \textbf{Mong Li Lee}
\\
\\
National University of Singapore
\\
 \small{
   {\tt zyan@u.nus.edu, \{peng.qi, whsu, leeml\}@nus.edu.sg}
 }
 \\
 {\small \href{https://yanzehong.github.io/trust-vl/}{\tt \color{pink}https://yanzehong.github.io/trust-vl}}
}

\begin{document}
\maketitle

\begin{abstract}
Multimodal misinformation, encompassing textual, visual, and cross-modal distortions, 
poses an increasing societal threat that is amplified by generative AI. Existing methods typically focus on a single type of distortion and struggle to generalize to unseen scenarios. 
In this work, we observe that different distortion types share common reasoning capabilities while also requiring task-specific skills. 
We hypothesize that joint training across distortion types facilitates knowledge sharing and enhances the model’s ability to generalize.
To this end, 
we introduce TRUST-VL, a unified and explainable vision-language model for general multimodal misinformation detection.
TRUST-VL incorporates a novel Question-Aware Visual Amplifier module, designed to extract task-specific visual features. 
  To support training, we also construct TRUST-Instruct, a large-scale instruction dataset containing 198K samples featuring structured reasoning chains aligned with human fact-checking workflows. 
  Extensive experiments on both in-domain and zero-shot benchmarks demonstrate that TRUST-VL achieves state-of-the-art performance, 
  while also offering strong generalization and interpretability. 
\end{abstract}

\section{Introduction}

\begin{figure}[t!]
  \centering
  \includegraphics[width=\linewidth]{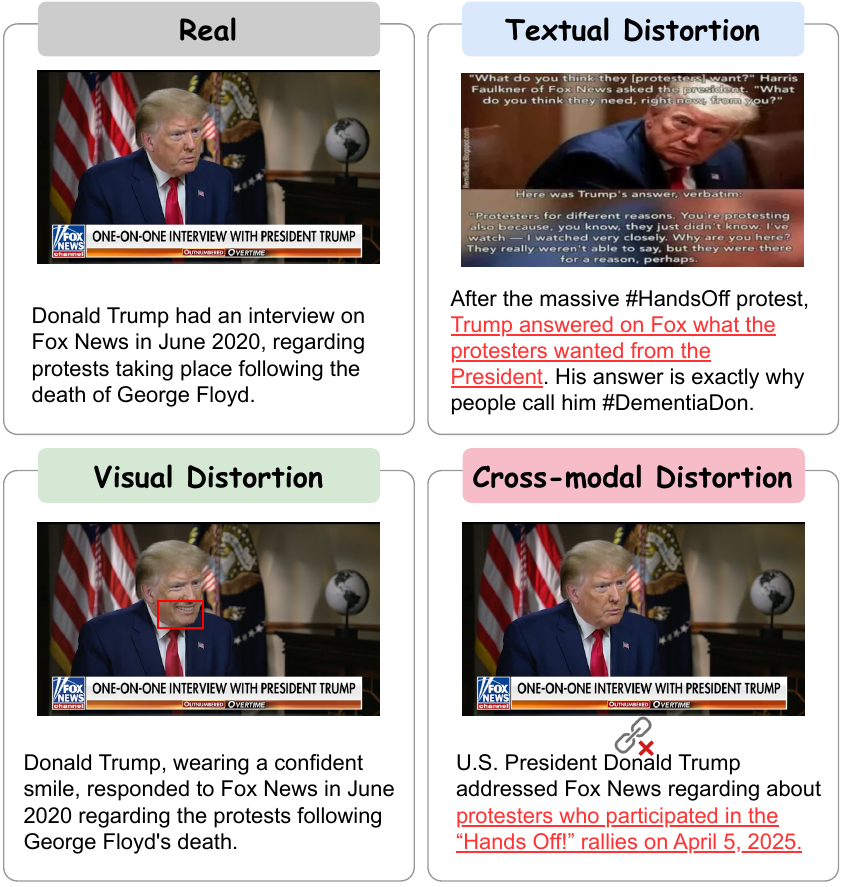}
  \caption{
  Examples of different distortion types in multimodal misinformation. 
  }
  \label{fig:intro_example}
\end{figure}

\begin{figure*}[t!]
  \centering
  \includegraphics[width=\linewidth]{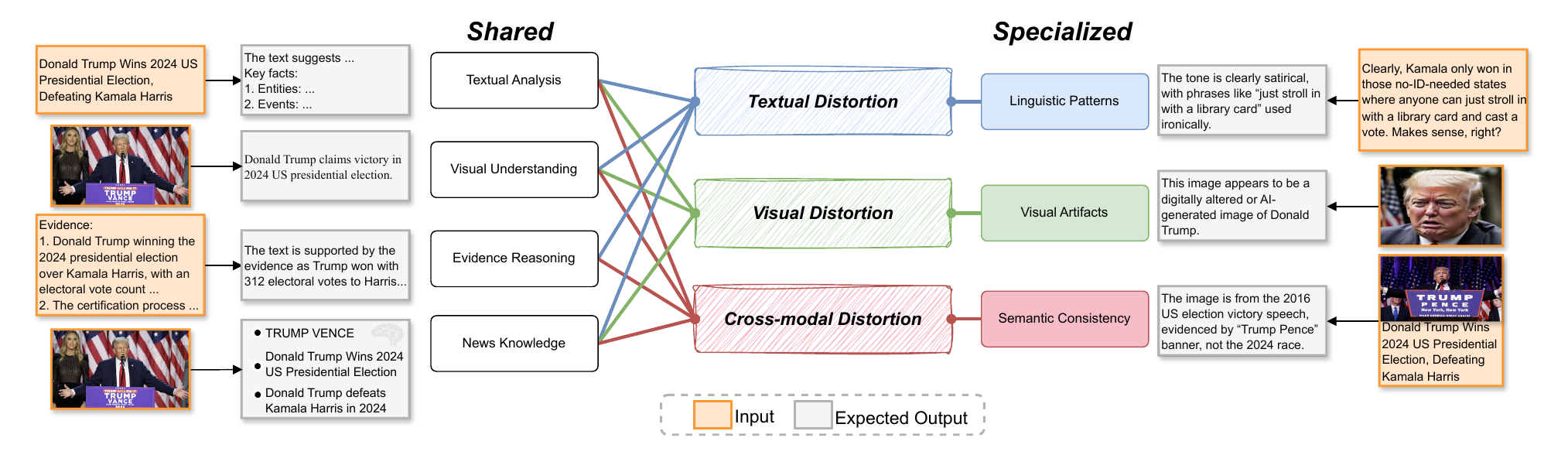}
  \caption{Overview of shared and specialized reasoning involved across misinformation detection tasks.}
  \label{fig:abilties}
\end{figure*}

Multimodal misinformation has become a fast-growing threat to society and has attracted wide attention in recent years. The rise of generative AI tools, while providing powerful capabilities for content creation, has also made it easier to produce misleading content and spread it at scale. For example, during the 2024 U.S. presidential election, foreign actors used AI-generated deepfakes and manipulated media to spread false narratives and influence voter perception, prompting official sanctions \cite{federspiel2023threats}. Therefore, it is urgent to develop automated methods to detect multimodal misinformation \cite{akhtar-etal-2023-multimodal,chen2024can, abdali2025multi}.

Multimodal misinformation is inherently a composite task, involving multiple sub-problems such as textual distortion, visual distortion, and cross-modal distortion. As illustrated in Figure~\ref{fig:intro_example}, {\it textual distortion} refers to discrepancies between the textual claim and the underlying facts, which can often be identified through linguistic patterns or textual entailment between the claim and retrieved evidence. {\it Visual distortion} involves tampered or AI-generated images, and can be detected by identifying subtle visual artifacts or inconsistencies. {\it Cross-modal distortion} (also known as out-of-context misinformation) arises when the image and text originate from different real-world events, which can be detected by assessing semantic consistency across modalities \cite{alam-etal-2022-survey,liu2025mmfakebench}.

Vision-language models (VLMs) have achieved impressive performance across a wide range of multimodal tasks
\cite{liu2023visual,dai2023instructblip,openai2024gpt4o,xue2024xgenmm,wang2024qwen2vl}.
Motivated by this, 
prior works have applied VLMs to 
 specific misinformation tasks such as fact checking \cite{yao2023mocheg,tahmasebi2024multimodal}, face manipulations \cite{liu2024fkaowl,huang2024ffaa}, and out-of-context detection \cite{qi2024sniffer}. 
 However, these models typically focus on a specific type of misinformation, and we empirically found that such
 single-task models often overfit and generalize poorly to unseen distortion types.

We observe that although detecting different distortion types requires \textit{specialized reasoning} (e.g., linguistic pattern recognition, visual artifact detection, and semantic consistency checks), they also rely on \textit{shared reasoning} (e.g.,  textual analysis, visual understanding, evidence-based reasoning, and familiarity with news knowledge) (see Figure ~\ref{fig:abilties}).
 For instance, multimodal content analysis is fundamental for in-depth reasoning, while evidence-based reasoning is crucial for tasks ranging from textual fact-checking to cross-modal inconsistency detection. 
 Motivated by this, we aim to build a unified framework that integrates both shared and specialized reasoning to effectively handle misinformation detection across diverse distortion types.

Developing a unified misinformation detection framework has several challenges: (1) Existing VLMs,  pretrained on general vision-language tasks, often lack sensitivity to subtle visual artifacts and cross-modal semantic inconsistency; (2) annotation standards vary widely across existing datasets, complicating unified learning \cite{thorne-etal-2018-fever,suryavardan2023factify2,liu2024fkaowl,luo2021newsclippings}; and 
(3) most datasets lack explicit reasoning annotations, and provide only binary or categorical labels without detailing the intermediate reasoning steps behind the veracity judgment, thus limiting a model’s ability to generate interpretable and persuasive explanations for  real-world fact-checking applications \cite{thibault2025guide,xu2023combating, akhtar-etal-2023-multimodal}.
These challenges highlight the need for new training paradigms with structured misinformation-specific reasoning annotations, along with comprehensive evaluation benchmarks to assess generalization across various misinformation tasks.

In this work, 
we observe that joint training across distortion types facilitates knowledge sharing and enhances the model's reasoning capabilities to generalize. Therefore, 
we introduce TRUST-Instruct, a large-scale dataset comprising reasoning-rich samples across diverse distortion types. 
Building upon this dataset, 
we propose TRUST-VL, a unified misinformation detection framework that enhances fine-grained visual understanding by conditioning perception on task-specific instructions.

Our main contributions can be summarized as follows:

    $\bullet$ We propose TRUST-VL, a unified and explainable vision-language model for general multimodal misinformation detection. It integrates a novel Question-Aware Visual Amplifier (QAVA) module to extract task-specific visual features and support reasoning across misinformation detection tasks.
    
    $\bullet$ We construct TRUST-Instruct, a large-scale instruction dataset of 198K samples with structured reasoning chains aligned with human fact-checking workflows, enabling effective joint training across diverse distortion types.
    
    $\bullet$ Extensive experiments on both in-domain and zero-shot benchmarks demonstrate that TRUST-VL achieves state-of-the-art performance, with superior generalization and interpretability compared to existing detectors and general VLMs.

\begin{figure*}[t!]
  \centering
  \includegraphics[width=0.8\linewidth]{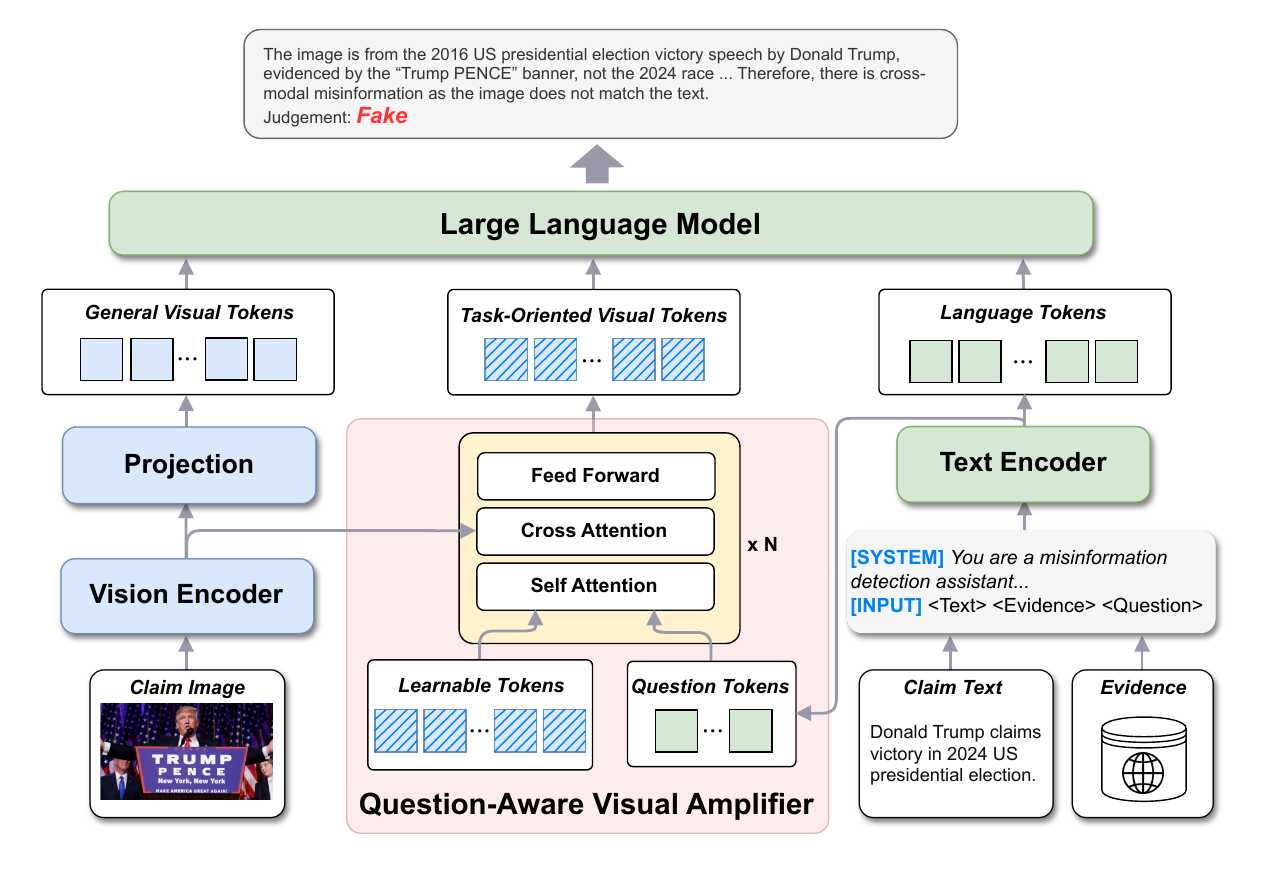}
  \caption{Architecture of TRUST-VL. Given an image-text pair and associated evidence, TRUST-VL first encodes multimodal inputs through vision and text encoders.  
  Other than projecting the visual features into general visual tokens, we also leverage the Question-Aware Visual Amplifier module, which utilizes a set of randomly initialized learnable tokens conditioned on task-oriented questions to generate task-oriented visual tokens.
  Finally, TRUST-VL outputs a structured and explainable detection judgment.}
  \label{fig:arch}
\end{figure*}

\section{Related Work}

Multimodal misinformation detection covers different sub-tasks that focus on different manipulation cues.
Works on \textit{textual distortion detection} use 
language models to fact check based on text only and often ignore the visual elements crucial for verifying many claims  \cite{thorne-etal-2018-fever,augenstein2019multifc,kotonya-toni-2020-explainable-automated,pan-etal-2023-fact}. 
For \textit{visual distortion detection}, recent efforts enhance VLMs with  forgery-aware reasoning and visual artifact localization by soft prompt tuning \cite{liu2024fkaowl} and instruction tuning \cite{li2024forgerygpt,huang2024ffaa}.
For \textit{cross-modal distortion detection},  \cite{tahmasebi2024multimodal, qi2024sniffer, xuan2024lemma} enhance VLM reasoning by introducing external evidence sources. Notably, SNIFFER \cite{qi2024sniffer} improves image-text consistency detection through a two-stage instruction tuning process.
However, these models are  trained on narrowly scoped misinformation types such as face swaps or hallucinated claims, and struggle to generalize to unseen types.

Recent studies have started exploring complex scenarios in which false information spans across modalities.
LRQ-FACT \cite{beigi2024lrqfact}  generates image- and text-focused questions using LLMs and VLMs, and synthesizes a final judgment through rule-based aggregation. \cite{liu2025mmfakebench} introduces MMD-Agent, a multi-agent framework that sequentially decomposes detection into textual, visual, and cross-modal subtasks, using step-wise prompting and retrieved evidence for improved reasoning. 
These multi-agent frameworks consist of loosely connected modules that are not jointly optimized for misinformation detection. In contrast, our proposed unified framework formulates misinformation tasks 
through a structured taxonomy of shared and specialized reasoning steps, and integrates them within a single VLM for end-to-end optimization and more effective detection.

\section{Proposed Framework}
\label{sec:method_trust-vl}
Our goal is to develop an explainable VLM for detecting multimodal misinformation with various types of distortions.
As illustrated in Figure~\ref{fig:arch}, the proposed TRUST-VL framework 
first retrieves relevant external evidence for the input image-text pair. The input text, evidence, and a task-specific question are encoded by a textual encoder, while the image is processed through a visual encoder equipped with a general projector and the Question-Aware Visual Amplifier. The resulting language and visual tokens are then jointly fed into an LLM to produce a final judgment with an explanation.

\subsection{TRUST-VL Model Architecture}

\begin{table*}[t]
\centering
\small
\begin{tabular}{p{3cm} p{11.5cm}}
\toprule
\textbf{Capabilities} & \textbf{Definitions} \\
\midrule
\multicolumn{2}{c}{\textcolor{gray!70}{\textit{\textbf{Shared Reasoning}}}} \\
\textbf{Textual Analysis} & Extracts key factual elements (e.g., entities, dates, events) from text and lists statements to be verified. \\
\textbf{Visual Understanding} & Interprets salient visual content (e.g., entities, scenes, actions) and identifies visual cues of manipulation, such as unnatural lighting, texture inconsistencies, distorted facial features, duplicated patterns, or incoherent backgrounds. \\
\textbf{Evidence Reasoning} & Cross-checks the claim against retrieved or user-provided evidence to identify factual support or contradiction. This capability is essential for verifying non-factual claims and detecting out-of-context image–text pairings. \\
\textbf{News Knowledge} & Recalls factual world knowledge about people, places, or events to contextualize the claim, even without using external information. \\
\midrule
\multicolumn{2}{c}{\textcolor{gray!70}{\textit{\textbf{Specialized Reasoning}}}} \\
\cellcolor{textualcolor}\textbf{Linguistic Patterns} &\cellcolor{textualcolor} Identifies rhetorical cues (e.g., bias, satire, sentiment) that may signal misleading or manipulative intent in the text. \\
\cellcolor{visualcolor}\textbf{Visual Artifacts} & \cellcolor{visualcolor} Detects pixel-level or visual artifacts (e.g., lighting issues, texture mismatches) indicating image manipulation or generation. \\
\cellcolor{crossmodalcolor}\textbf{Semantic Consistency} & \cellcolor{crossmodalcolor} Assesses the semantic matching between textual and visual modalities to detect out-of-context misinformation. Discrepancies can indicate that authentic images are being misused to support misleading narratives. \\
\bottomrule
\end{tabular}
\caption{Taxonomy of reasoning capabilities required for multimodal misinformation detection.}
\label{tab:taxonomy-capabilities}
\end{table*}

\noindent\textbf{Model Input.}
Given a multimodal claim consisting of an image $C_I$ and associated text $C_T$, TRUST-VL first retrieves external evidence from the open-domain web through the cross-modal retrieval \cite{abdelnabi2022open}.  Specifically, we retrieve the top-$m$ most relevant direct evidence ($E_{1:m}^{dir}$) using an image retriever guided by $C_T$, which is converted into captions via image-to-text generation. At the same time, we retrieve the top-n most relevant inverse evidence ($E_{1:n}^{inv}$) using a text retriever queried by $C_I$. Additionally, TRUST-VL incorporates context evidence ($E_{1:k}^{ctx}$), such as Wikipedia articles or expert annotations, provided either by users or downstream benchmarks.

\noindent\textbf{Base VLM. }
We follow the architecture of LLaVA \cite{liu2023visual} to build our own explainable VLM for multimodal misinformation detection. 
Besides the pretrained LLM and visual encoder, 
we use lightweight MLP projectors \cite{liu2023visual,liu2024improved} to connect image features to the word embedding space of the language model and then fine-tune the model on instruction-formatted datasets to improve generalization and controllability.

\noindent\textbf{Question-Aware Vision Amplifier.} Existing VLMs  typically rely on high-level semantic cues (scene, context, or objects) to detect visual distortions such as face manipulation. However, they often struggle  with subtle manipulations that modify facial expressions while preserving identity. 
Directly incorporating such visual distortions ~\cite{luo2021generalizing,li2021frequency,liu2024fkaowl} may degrade the model's performance on other types of distortions, due to potential overfitting to specific visual artifacts or a shift in representation focus.

\begin{figure*}[t!]
  \centering
  \includegraphics[width=\linewidth]{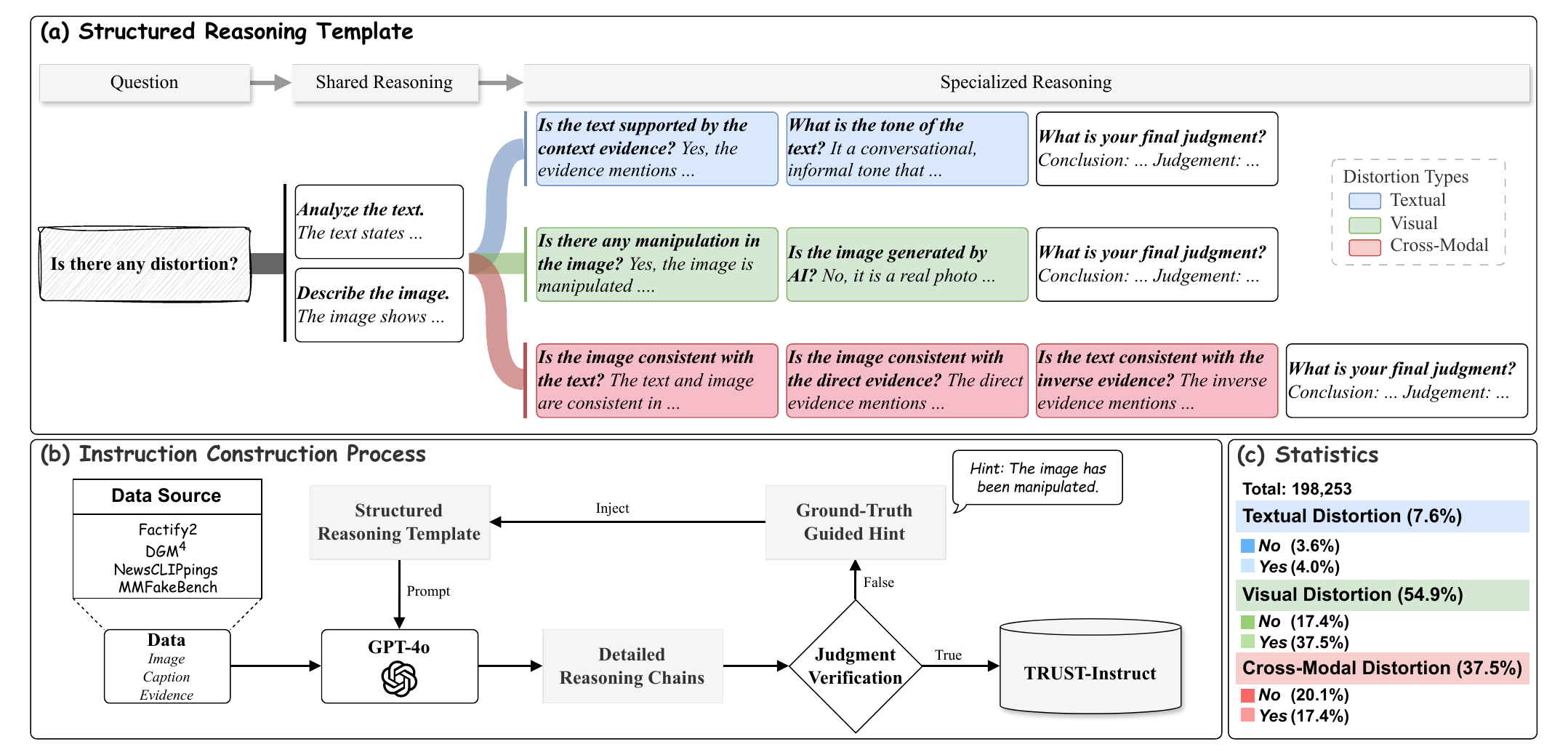}
  \caption{Construction of TRUST-Instruct using structured reasoning template. TRUST-Instruct comprises 198K diverse samples spanning various distortion types, each annotated with rich, step-by-step reasoning chains. 
  }
  \label{fig:trust_instruct}
\end{figure*}

To overcome this limitation, we introduce the Question-Aware Vision Amplifier (QAVA), a novel module inspired by the Q-Former \cite{li2023blip2,dai2023instructblip}. Unlike previous methods that rely solely on textual instructions, which often introduce irrelevant cues, QAVA employs learnable tokens conditioned specifically on explicit, task-specific question templates corresponding to different distortion types.
Within QAVA, these tokens first utilize self-attention to capture the question context and then apply cross-attention to the image features to extract precise, task-relevant visual cues. The resulting enhanced visual representations serve as soft visual prompts for the LLM,  guiding its reasoning process and thus improving the detection accuracy, especially for subtle visual distortions.

\subsection{Construction of TRUST-Instruct}
\label{sec:method_trust_instruct}

We construct an instruction dataset to enhance reasoning capabilities of TRUST-VL. These 
 capabilities can be grouped into \emph{shared} and \emph{specialized} reasoning as shown in Table~\ref{tab:taxonomy-capabilities}.
  These capabilities guide the construction of our TRUST-Instruct dataset, each addressing characteristic misinformation patterns spanning text, vision, and cross-modal reasoning steps (see Figure~\ref{fig:trust_instruct}).

\noindent\textbf{Structured Reasoning Template.} 
We mimic the human fact-checking process \cite{vlachos-riedel-2014-fact,warren2025show}
and regard misinformation detection as a sequence of reasoning steps tailored to different categories of distortions.  We design specific sub-queries that guide the model through a structured, step-by-step verification process for each distortion type.

This verification process consists of common shared reasoning steps for analyzing the text and describing the image across all distortion types, before branching into 
task-specific reasoning. For textual distortion reasoning, we evaluate the tone, stance, and evidence support. For 
visual distortion reasoning, we focus on manipulated artifacts or AI-generated patterns. For
cross-modal distortion reasoning, we verify the semantic consistency between image, caption, and retrieved evidence.
This structured reasoning approach mirrors real-world fact-checking workflows and provides an interpretable, robust detection judgment.

\noindent\textbf{Instruction Generation.} 
Motivated by the success of  generative models in automated instruction generation \cite{zhang2024multimodal}, we propose a structured  pipeline to construct reasoning instructions (see Figure~\ref{fig:trust_instruct}(b)).
To  create a  comprehensive dataset covering multiple  distortion types, we curate a collection of <text, image, ground-truth label> triplets from several established datasets: Factify2 \cite{suryavardan2023factify2} for textual claims with and without distortion; DGM\textsuperscript{4} \cite{shao2023dgm} for visual manipulations (e.g., face swaps and face-attribute editing) alongside their authentic counterparts; MMFakeBench \cite{liu2024mmfakebench} for  visual forgeries that are AI-generated or Photoshop-edited; and NewsCLIPpings \cite{luo2021newsclippings} for out-of-context image–text mismatches.

Based on this collection, we generate the  instruction data by providing the multimodal input claims and their associated evidence to GPT-4o \cite{openai2024gpt4o}, which is prompted with a carefully designed reasoning template to produce detailed reasoning chains for misinformation detection. Each generated chain undergoes a rigorous verification stage to ensure consistency with the ground-truth labels. When inconsistencies are detected, the prompts are iteratively refined with data-driven hints based on the ground truth to  guide GPT-4o toward accurate reasoning outputs.

To ensure the quality of TRUST-Instruct, we manually inspect a subset of the generated instructions to verify that: (1) the generated instructions and reasoning chains are coherent and align with the distortion type; (2) the task-specific reasoning steps are carried out only after the shared reasoning steps have been  completed; (3) the task-specific (specialized) reasoning steps are correct; and (4) the final veracity labels match the  ground truth. 
98.5\% of the generated instructions pass our inspection and we filter out the remaining ones that fail to meet these criteria.
 The final TRUST-Instruct dataset comprises 198,253 high-quality instructions spanning three distortions (see Figure~\ref{fig:trust_instruct}(c)).

\begin{figure*}[t!]
  \centering
  \includegraphics[width=\linewidth]{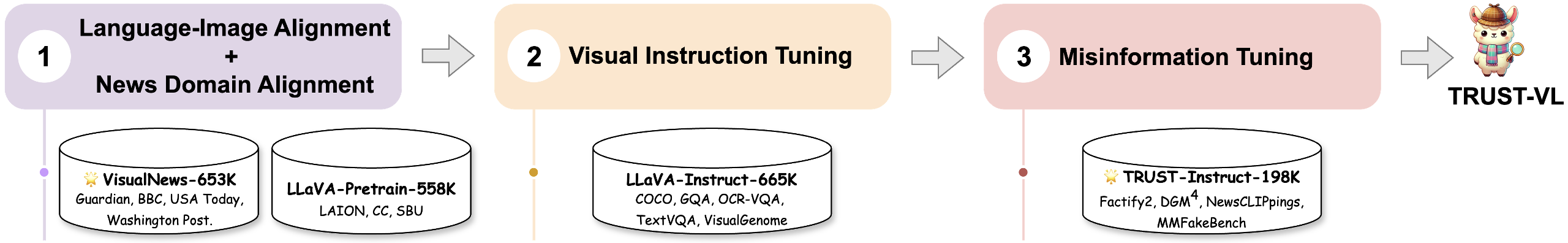}
  \caption{Progressive training strategy.}
  \label{fig:training_stages}
\end{figure*}

\begin{table*}[t!]
\small
\centering
\resizebox{\textwidth}{!}{
\begin{tabular}{lccccccc}
\toprule
\multirow{2}{*}{\textbf{Dataset}} & \multicolumn{4}{c}{\textbf{In-Domain}} & \multicolumn{3}{c}{\textbf{Out-of-Domain}} \\
\cmidrule(lr){2-5} \cmidrule(lr){6-8}
& \textbf{MMFakeBench} & \textbf{Factify2} & \textbf{DGM\textsuperscript{4}-Face} & \textbf{NewsCLIPpings} & \textbf{MOCHEG} & \textbf{Fakeddit-M} & \textbf{VERITE}\\
\toprule
Distortion Types        & Mixed & Textual & Visual & Cross-modal & Textual & Visual & Cross-modal \\
\# Label: \textit{No} & 300 & 1500 & 467 & 3632 & 200 & 200 & 200 \\
\# Label: \textit{Yes} & 700 & 1500 & 433 & 3632 & 200 & 200 & 200 \\

\bottomrule
\end{tabular}
}
\caption{Evaluation Dataset Distribution.}
\label{tab:dataset_distribution}
\end{table*}

\subsection{Training of TRUST-VL Model} 
Figure~\ref{fig:training_stages} shows the 
three-stage 
training process that progressively enhance the capabilities of our TRUST-VL model. 

\noindent\textbf{Stage 1.} We begin by training the projection module for one epoch on 1.2 million image–text pairs (653K news samples from VisualNews \cite{liu2020visualnews} and 558K samples from the LLaVA training corpus \cite{liu2024improved}). This stage aligns the visual features with the language model. 

\noindent\textbf{Stage 2.} Next, we jointly train  the LLM and the projection module for one epoch
using 665K synthetic conversation samples from the LLaVA training corpus \cite{liu2024improved} to improve the  model’s ability to follow complex instructions. 

\noindent\textbf{Stage 3.} Finally, we fine-tune the full model on 198K reasoning samples from TRUST-Instruct for three epochs to further enhance its misinformation-specific reasoning capabilities.

\section{Performance Study}

\noindent\textbf{Datasets.} 
To demonstrate the generalization capability of TRUST-VL, we evaluate the model on a diverse collection of in-domain and out-of-domain 
datasets covering textual, visual, and cross-modal distortions (see Table~\ref{tab:dataset_distribution}). 
{\it In-domain datasets} include MMFakeBench \cite{liu2025mmfakebench} which has mixed distortion types; Factify2 \cite{suryavardan2023factify2}, a fact-checking benchmark for multimodal claim verification; DGM\textsuperscript{4}-Face \cite{shao2023dgm}, focused on detecting deepfake-powered facial manipulations such as face swaps; and NewsCLIPpings \cite{luo2021newsclippings}, the largest synthetic benchmark for out-of-context (OOC) misinformation detection, created by  replacing the images in original claims with  semantically related but event-mismatched images. 
{\it Out-of-domain datasets} include MOCHEG \cite{yao2023mocheg}, a textual misinformation dataset with journalist-verified claims; Fakeddit-M \cite{nakamura2020fakeddit}, a Reddit-sourced visual distortion dataset under the Manipulated Content category (e.g.,  digitally edited images); and VERITE \cite{papadopoulos2024verite}, a real-world OOC benchmark with modality-balanced image-text pairs. 

\noindent\textbf{Baselines.}
We compare TRUST-VL with both general-purpose VLMs and specialized misinformation detectors. For {\it general-purpose VLMs}, we include BLIP-2 \cite{li2023blip2}, InstructBLIP \cite{dai2023instructblip}, LLaVA \cite{liu2023visual}, LLaVA-NeXT \cite{li2024llavanext}, xGen-MM \cite{xue2024xgenmm}, and Qwen2-VL \cite{wang2024qwen2vl}, which are all open-source VLMs primarily designed for multimodal understanding and reasoning tasks. 
We also include GPT-4o \cite{openai2024gpt4o} and o1 \cite{openai2024o1}, two advanced closed-source VLMs.
For {\it specialized misinformation detectors}, we consider SNIFFER \cite{qi2024sniffer}, an explainable VLM-based detector for OOC misinformation through a two-stage instruction; 
MMD-Agent \cite{liu2025mmfakebench}, a multi-agent framework that utilizes VLMs for three sequential stages of veracity checking, and LRQ-FACT \cite{beigi2024lrqfact}, a fact-checking system based on a multi-LLM architecture that improves context reasoning.

\noindent\textbf{Implementation Details.} 
We use LLaVA-1.5 \cite{liu2024improved} with \verb|vicuna-13b-v1.5| as the LLM and CLIP (\verb|ViT-L/14|) as the image encoder. The learning rates are set to 2e-5 for the LLM and 2e-6 for the vision encoder, with a batch size of 128. 
All models are trained 
on 8 Nvidia H100 (80G) GPUs. 
We evaluate model performance using  accuracy (Acc.) and macro-F1.

\subsection{Performance Comparison}

\begin{table*}[t!]
    \centering
    \small
    \resizebox{\textwidth}{!}{
    \begin{NiceTabular}{lccccccccccccccc}
        \CodeBefore
            \rectanglecolor{skyblue!30}{17-1}{17-18}
            \rectanglecolor{mygreen}{18-1}{18-19}
        \Body
        \toprule
        \multirow{3}{*}{\textbf{Methods}} & \multirow{3}{*}{\textbf{Avg. Acc.}} 
        & \multicolumn{8}{c}{\textbf{In-Domain}} 
        & \multicolumn{6}{c}{\textbf{Out-of-Domain}} \\
        \cmidrule(lr){3-10} \cmidrule(lr){11-16}
        & & \multicolumn{2}{c}{\textbf{MMFakeBench}} & \multicolumn{2}{c}{\textbf{Factify2}} & \multicolumn{2}{c}{\textbf{DGM$^4$-Face}} & \multicolumn{2}{c}{\textbf{NewsCLIPpings}} 
        & \multicolumn{2}{c}{\textbf{MOCHEG}} & \multicolumn{2}{c}{\textbf{Fakeddit-M}} & \multicolumn{2}{c}{\textbf{VERITE}} \\
        \cmidrule(lr){3-4} \cmidrule(lr){5-6} \cmidrule(lr){7-8} \cmidrule(lr){9-10}
        \cmidrule(lr){11-12} \cmidrule(lr){13-14} \cmidrule(lr){15-16}
        & & Acc. & F1 & Acc. & F1 & Acc. & F1 & Acc. & F1 & Acc. & F1 & Acc. & F1 & Acc. & F1 \\
        \midrule
        \multicolumn{16}{l}{\textcolor{gray!80}{\textit{\textbf{General-purpose VLMs}}}} \\
        BLIP2 & 53.36 & 37.40 & 34.45 & 54.30 & 42.38 & 47.70 & 34.35 & 50.14 & 34.28 & 62.50 & 57.16 & 70.75 & 70.19 & 50.75 & 37.35 \\
        InstructBLIP & 58.41 & 57.30 & 56.38 & 66.83 & 66.48 & 50.40 & 48.66 & 53.85 & 50.71 & 63.25 & 60.85 & 64.75 & 62.83 & 52.50 & 49.60 \\
        LLaVA & 60.25 & 62.60 & 61.72 & 79.59 & 79.10 & 46.41 & 38.14 & 45.87 & 48.54 & 66.50 & 64.71 & 68.00 & 66.67 & 52.75 & 49.80 \\
        xGen-MM & 62.20 & 65.40 & 62.77 & 86.03 & 86.04 & 50.10 & 49.68 & 59.87 & 59.18 & 59.50 & 56.32 & 60.00 & 53.45 & 54.50 & 54.41 \\
        LLaVA-NeXT & 62.35 & 71.60 & 65.99 & 79.60 & 79.09 & 53.40 & \underline{52.21} & 59.86 & 59.37 & 58.25 & 52.52 & 59.00 & 52.36 & 54.75 & 54.57 \\
        Qwen2-VL & 69.85 & 67.00 & 66.28 & 89.40 & 89.37 & 48.10 & 41.63 & 70.94 & 69.91 & 66.25 & 64.57 & \underline{77.25} & \underline{76.96} & 70.00 & 68.94 \\
        GPT-4o & 76.16 & 83.10 & 80.88 & 88.37 & 88.21 & \underline{57.14} & 49.24 & 86.51 & 86.51 & 77.00 & 76.81 & 73.50 & 73.12 & 67.50 & 67.57 \\
        o1 & \underline{77.74} & \underline{83.90} & \underline{82.41} & \underline{96.90} & \underline{96.90} & 50.06 & 38.06 & 86.80 & 86.54 & \underline{81.50} & \underline{81.38} & 73.25 & 73.07 & 71.75 & 71.66 \\
        \midrule
        \multicolumn{16}{l}{\textcolor{gray!80}{\textit{\textbf{Misinformation Detectors}}}} \\
        MMD-Agent & 56.11 & 69.10 & 48.68 & 71.03 & 69.35 & 48.30 & 48.29 & 53.06 & 41.12 & 54.25 & 43.72 & 42.25 & 42.24 & 54.75 & 47.00 \\
        SNIFFER & 61.17&51.40&51.33&61.00&55.97&47.20&37.96&\underline{88.85}&\underline{88.85}&53.75&50.73&53.50&51.13&\underline{72.50}&\underline{72.02}\\
        LRQ-FACT & 66.60 & 71.30 & 74.00 & 86.63 & 89.79 & 41.80 & 44.14 & 68.19 & 73.45 & 66.25 & 69.25 & 67.25 & 71.77 & 64.75 & 68.32 \\
        \textbf{TRUST-VL} & \textbf{86.16} & \textbf{87.30} & \textbf{85.42} & \textbf{99.50} & \textbf{99.50} & \textbf{88.50} & \textbf{88.39} & \textbf{90.35} & \textbf{90.35} & \textbf{82.75} & \textbf{82.58} & \textbf{82.50} & \textbf{82.20} & \textbf{73.75} & \textbf{73.61} \\
        $\Delta$& ↑8.42 & ↑3.40 & ↑3.01 & ↑2.60 & ↑2.60 & ↑31.36 & ↑36.18 & ↑1.50 & ↑1.50 & ↑1.25 & ↑1.20 & ↑5.25 & ↑5.24 & ↑1.25 & ↑1.59 \\
        \bottomrule
    \end{NiceTabular}
    }
    \caption{Performance (\%)  comparison between TRUST-VL and other baseline VLMs across in-domain and out-of-domain datasets. The best score is highlighted in blue, and the second-best score is underlined. The absolute improvement over the second-best model is highlighted in green.}
    \label{tab:main_results}
\end{table*}

\begin{table}[t!]
\small
\centering
\resizebox{\columnwidth}{!}{
\begin{tabular}{llc}
\toprule
\textbf{Dataset} & \textbf{Model} & \textbf{Acc.} \\
\midrule
\multirow{2}{*}{Factify2} 
& LVLM4FV \cite{tahmasebi2024multimodal} & 80.13 \\
& TRUST-VL & 99.50 \\
\midrule
\multirow{2}{*}{DGM$^{4}$-All} 
& HAMMER \cite{shao2023dgm} & 86.39 \\
& TRUST-VL & 87.26 \\
\midrule
\multirow{2}{*}{NewsCLIPpings} 
& SNIFFER \cite{qi2024sniffer} & 88.85 \\
& TRUST-VL & 90.35 \\
\bottomrule
\end{tabular}
}
\caption{Performance (\%) comparison with task-specific baselines across representative datasets.}
\label{tab:task_specific_baselines}
\vspace{-0.5cm}
\end{table}

Table \ref{tab:main_results} shows the results. We see that:
\begin{itemize}
[itemsep=.8pt, topsep=2pt, leftmargin=1em]
    \item Our proposed TRUST-VL significantly outperforms all baselines on both in-domain and out-of-domain datasets, achieving more than 8 percentage points improvement in average accuracy. This demonstrates that our proposed TRUST-VL effectively captures the key detection cues across different distortion types and generalizes well to unseen news claims. 

    \item General-purpose VLMs, particularly OpenAI-o1, exhibit competitive performance on textual and cross-modal distortions, but  struggle with subtle visual manipulations. Specifically, o1 achieves an overall accuracy of 77.74\%, but its performance drops significantly on DGM\textsuperscript{4}-Face (50.06\%), indicating challenges in detecting manipulated facial content. Besides, o1 also outperforms GPT-4o, especially on textual distortions, suggesting that the enhanced reasoning capabilities can benefit misinformation detection.
    \item Existing multimodal misinformation detectors that rely on multiple independent LLMs for step-by-step reasoning perform worse than general-purpose VLMs.  MMD-Agent and LRQ-FACT achieve average accuracies of 56.11\% and 66.60\%, respectively. This may be due to conflicting reasoning paths across  modules, which undermine the overall decision-making process.
\end{itemize}

\noindent\textbf{Comparison with Task-Specific Models.} To further demonstrate the effectiveness of our unified framework, we  compare TRUST-VL against strong task-specific baselines on representative benchmarks.  Table~\ref{tab:task_specific_baselines} shows that  our unified approach not only generalizes across diverse distortion types but also achieves superior performance compared to specialized models. For Factify2, the primary challenges stem from its long textual context and the need for evidence reasoning. We attribute the strong performance of TRUST-VL to the advanced capabilities of the underlying large language model in handling complex reasoning in text modality.

\begin{table*}[t!]
\centering
\small
\begin{tabular}{lcccccccc}
\toprule
\multirow{2}{*}{\textbf{Variants}} & \multicolumn{2}{c}{\textbf{MMFakeBench}} & \multicolumn{2}{c}{\textbf{Factify2}} & \multicolumn{2}{c}{\textbf{DGM\textsuperscript{4}-Face}} & \multicolumn{2}{c}{\textbf{NewsCLIPpings}} \\
\cmidrule(lr){2-3} \cmidrule(lr){4-5} \cmidrule(lr){6-7} \cmidrule(lr){8-9}
 & \textbf{Acc.} & \textbf{F1} & \textbf{Acc.} & \textbf{F1} & \textbf{Acc.} & \textbf{F1} & \textbf{Acc.} & \textbf{F1} \\
\midrule
TRUST-VL-13B & \textbf{87.30} & \textbf{85.42} &  \textbf{99.50} & \textbf{99.50} & \textbf{88.50} & \textbf{88.39} & \textbf{90.35} & \textbf{90.35}\\
w/o Reasoning & 83.60 & 81.25 & 87.31 & 87.30 & 80.00 & 79.91 & 85.99 & 85.98 \\
w/o Common Reasoning & 84.60 & 81.42 & 99.20 & 99.20 & 70.90 & 70.68 & 89.00 & 89.00 \\
w/o QAVA & 84.60&82.16&89.17 & 89.17 & 72.79 & 72.59 & 87.31 & 87.30 \\
LLM Size: 7B & 85.90 & 83.65 &  99.33 & 99.33 & 80.90 & 80.64 & 88.79 & 88.79\\
\bottomrule

\end{tabular}

\caption{Ablation study of different components in TRUST-VL. 
}
\label{tab:ablation}
\end{table*}

\subsection{Ablation Study}
We conduct ablation studies to evaluate the effect of different components in TRUST-VL, joint training across distortions and QAVA token count. 

\noindent{\bf Effect of Model Components. }
We implement the following variants of TRUST-VL: (a) {\it w/o Reasoning} where the model is trained only for binary classification ({\it i.e.}, real vs. fake), without generating structured reasoning chains; (b)
{\it w/o Common Reasoning} where the shared reasoning steps (text analysis and visual understanding) are removed during instruction data construction; (c) {\it w/o QAVA} where  QAVA module is removed from the model.
Table~\ref{tab:ablation} shows the results. We observe that:
\begin{itemize}
[itemsep=.8pt, topsep=2pt, leftmargin=1em]

\item TRUST-VL {\it w/o Reasoning} leads to substantial performance degradation (4–12 percentage points across datasets), highlighting the importance of structured reasoning supervision for accurate judgment. 

\item TRUST-VL {\it w/o Common Reasoning}  results in a noticeable performance decline, particularly on datasets involving fine-grained visual manipulations. This suggests that textual and visual descriptions provide crucial semantic grounding for subtle distortion detection.

\begin{figure}[t!]
  \centering
  \includegraphics[width=0.95\linewidth]{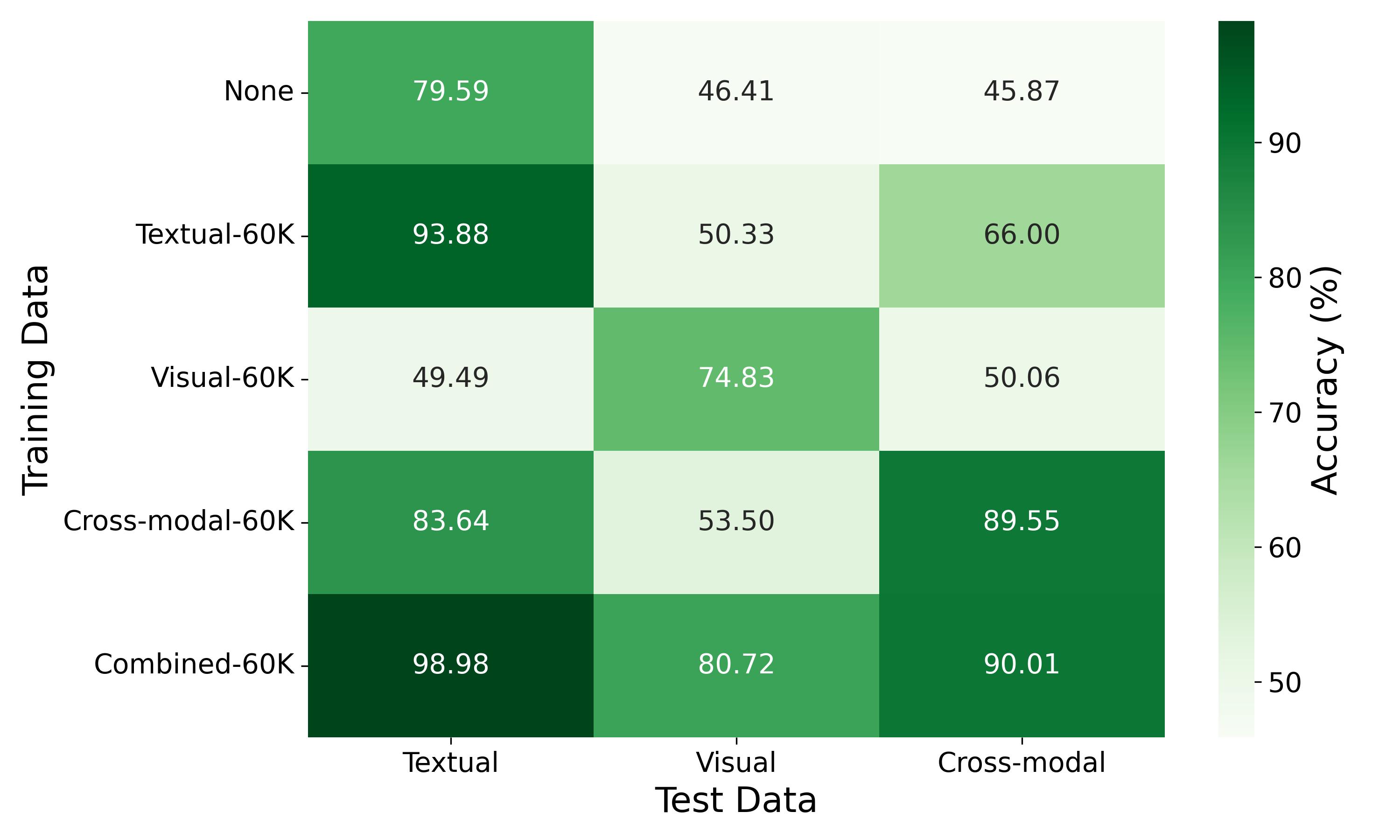}
  \caption{
  Accuracy heatmap of LLaVA across different training and testing distortion types.
  The first row (``None'') refers to the performance of the original LLaVA baseline without any training.
  }
  \label{fig:preliminary-results}
\end{figure}

\item TRUST-VL {\it w/o QAVA} results in a performance drop across all datasets, with the largest degradation of 15.71\%  on visual distortion tasks. This confirms the effectiveness of QAVA in learning task-specific visual representations.

\end{itemize}

\noindent{\bf Effect of Backbone Model Size. }
We replace the 13B backbone LLM with a smaller 7B version. Table~\ref{tab:ablation} shows that while using a 7B LLM leads to a  moderate performance decline, it still outperforms the second-best baseline reported in  Table~\ref{tab:main_results}. This highlights   the robustness and efficiency of our proposed framework and instruction data, even when smaller backbone models are used.

\noindent{\bf Effect of Joint Training. } \label{sec:methodology_preliminary}
To examine whether different distortion types  benefit from joint training, we conduct a small-scale experiment based on the original LLaVA model. We separately train the model using instruction data from each individual distortion type (textual, visual, or cross-modal), and compare the results with a jointly trained model using a balanced mix of all three types. For fair comparison, all models are trained on 60K samples.
As shown in Figure~\ref{fig:preliminary-results}, models trained on a single distortion type generally perform well on in-domain evaluation but struggle to generalize to unseen distortions. In contrast, the jointly trained model achieves consistently better performance across all distortion types, confirming that shared reasoning capabilities can be enhanced through joint training and transferred across tasks.

\begin{figure}[t!]
    \centering
    \includegraphics[width=0.95\linewidth]{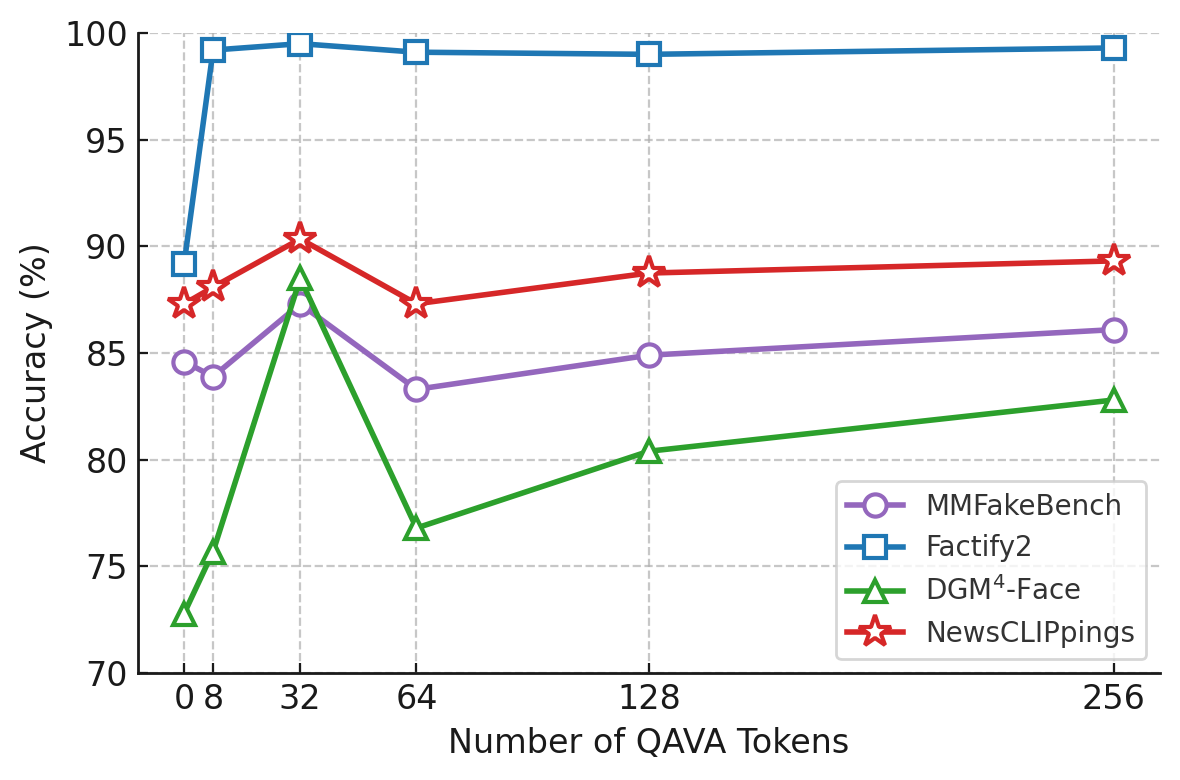}
    \caption{The impact of different numbers of learnable QAVA tokens across datasets.}
    \label{fig:exp_ablation_qava}
\end{figure}

\noindent{\bf Effect of QAVA Token Count. }
We also examine how the number of learnable visual tokens in the QAVA module influences the performance of TRUST-VL. Figure~\ref{fig:exp_ablation_qava} shows that  the QAVA module consistently improves accuracy across all datasets, with  notable gains on DGM\textsuperscript{4}-Face (accuracy increases from 72.79\% to 88.50\%), showing its critical role in detecting visual distortions.   Increasing the QAVA token count initially leads to performance gains, but beyond a certain point, further increases yield diminishing or even negative returns. Specifically, 32 tokens achieve the best performance across all datasets, suggesting they provide an optimal balance—sufficient to capture task-specific visual differences while avoiding excessive computational overhead and the risk of overfitting.

\begin{figure*}[t!]
  \centering
  \includegraphics[width=\linewidth]{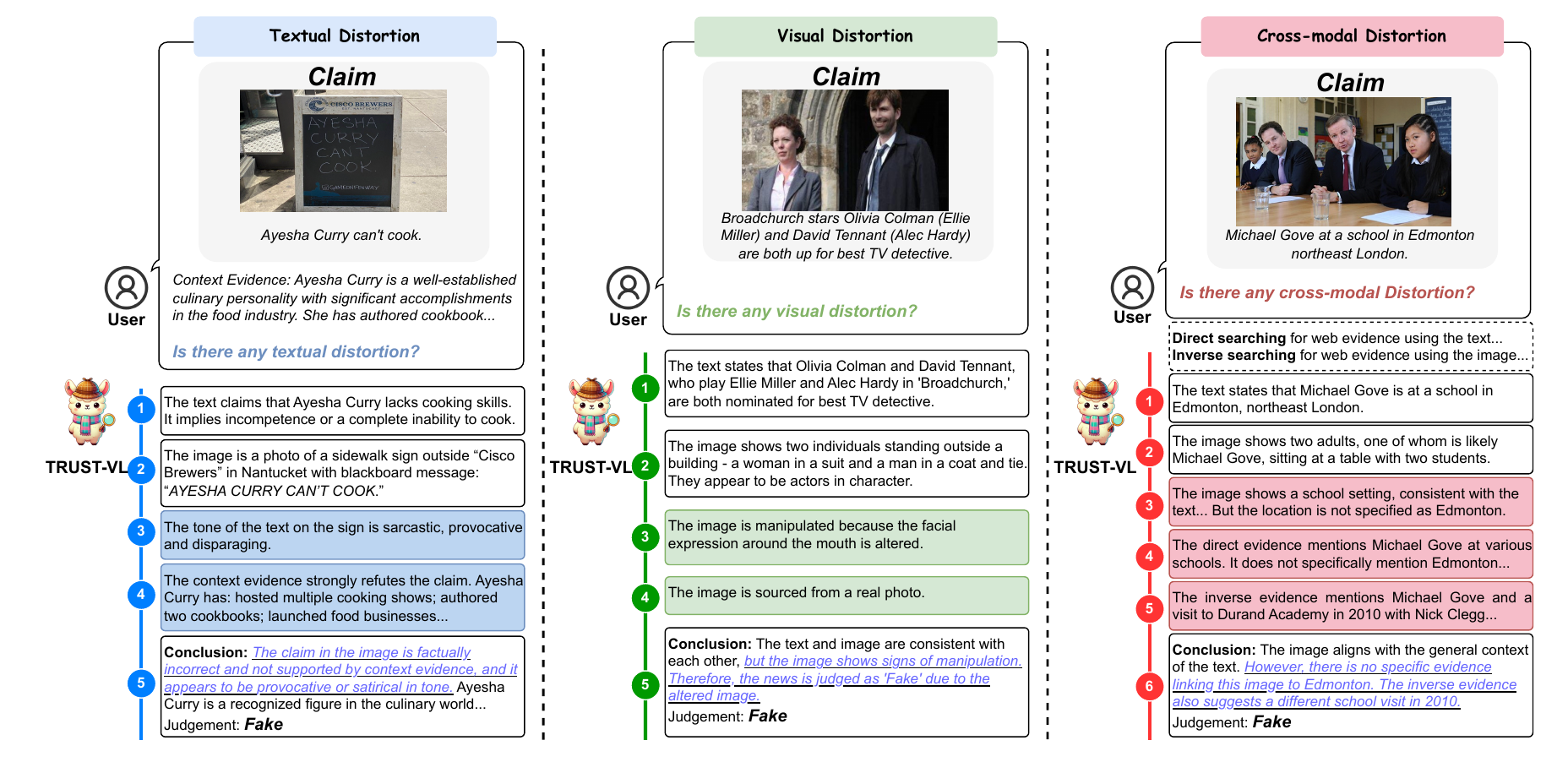}
  \caption{Example multimodal distortion spanning textual, visual, and cross-modal scenarios. }
  \label{fig:case_1}
  
\end{figure*}

\begin{figure*}[t!]
  \centering
  \includegraphics[width=0.95\linewidth]{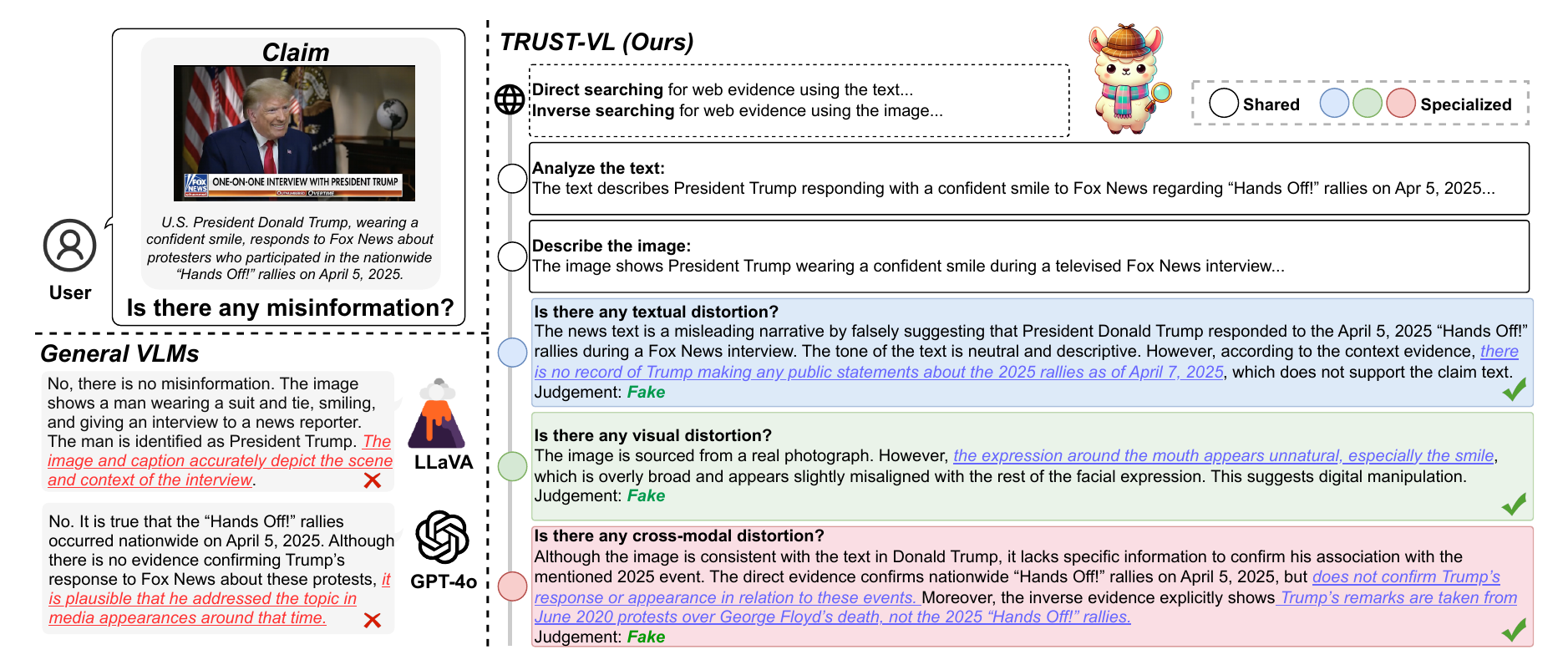}
  \caption{Comparison between  TRUST-VL and general large vision-language models on a complex case where false information spans across multiple modalities at the same time.\\}
  \label{fig:case_2}
\end{figure*}

\subsection{Case Study}

Figure~\ref{fig:case_1} shows three cases
that both general-purpose VLMs and specialized detectors  fail to handle. 
In contrast, TRUST-VL  correctly identifies all cases  with a structured sequence of reasoning steps. The first case involves the textual claim \q{Ayesha Curry can’t cook,} which contradicts well-documented facts and is presented in a satirical tone  likely to mislead users.
The second features manipulated photos of actors Olivia Colman and David Tennant, where subtle alterations to their facial expressions  convey deceptive emotions. The third pairs
an authentic image of politician Michael Gove  with an incorrect caption, producing a cross-modal mismatch.
These cases highlight the need for robust detection capable of addressing multiple, simultaneous distortions across text and images.

Figure~\ref{fig:case_2} shows a case 
where general VLMs fail to detect visual distortions on the person’s face, as well as cross-modal distortion (\textit{i.e.,} event mismatch between the text and image). General-purpose models like GPT-4o and LLaVA overlook these subtle manipulations and accept the content as factual. In contrast, TRUST-VL accurately identifies the misinformation by conducting multi-step reasoning, cross-referencing temporal and contextual evidence, and pinpointing inconsistencies across modalities. This demonstrates TRUST-VL’s superior ability to handle nuanced, real-world misinformation scenarios that require both shared and task-specific reasoning capabilities.

\section{Conclusion}
In this work, we tackle the challenge of multimodal misinformation detection involving textual, visual, and cross-modal distortions.
Recognizing that these tasks share common reasoning capabilities while also requiring specialized skills for each distortion type,   we propose joint training across distortion types to enhance model performance.
We introduce TRUST-VL, a unified, explainable VLM  with a novel Question-Aware Visual Amplifier module. We also construct the TRUST-Instruct dataset with structured reasoning chains that mimic human fact-checking.
Extensive experiments show that TRUST-VL achieves state-of-the-art results on both in-domain and out-of-domain benchmarks.

\section*{Acknowledgments}

This work is supported by the Ministry of Education, Singapore, under its MOE AcRF Tier 3 Grant (MOE-MOET32022-0001).

\section*{Limitations}
Although TRUST-VL achieves strong performance, it has several limitations.
First, the structured reasoning chains 
are guided by manually designed task queries, rather than being learned or evolved by the model. Incorporating reinforcement learning could further enhance the adaptability of the reasoning process.
Second, while visual evidence is retrieved, it is converted to text for reasoning. The more direct comparison in the visual space could offer richer signals. Lastly, our focus on visual distortion is limited to face-related manipulations, leaving other forms such as object-based or video misinformation for future exploration.

\bibliography{main}

\appendix

\section{Model Details}

\begin{table*}[t!]
\centering
 \small
\begin{tabular}{ll}
\hline
\textbf{Configurations} & \textbf{Details} \\ \hline
\textbf{Architecture} & 
\begin{tabular}[c]{@{}l@{}}
\textbf{Image Encoder}: CLIP-Large (336×336) \\ 
\textbf{Projector}: 2-Layer MLP \\ 
\textbf{QAVA}: 6 Transformer Layers with 32 Learnable Tokens \\ 
\textbf{LLM}: Vicuna-1.5 13B
\end{tabular} \\ \hline
\textbf{\# Total Parameters} & 13B \\ \hline
\textbf{Stage-1} & 
\begin{tabular}[c]{@{}l@{}}
\textbf{Training Data}: 1211K \\ 
\textbf{Trainable Module}: Projector
\end{tabular} \\ \hline
\textbf{Stage-2} & 
\begin{tabular}[c]{@{}l@{}}
\textbf{Training Data}: 665K \\ 
\textbf{Trainable Module}: LLM, Projector
\end{tabular} \\ \hline
\textbf{Stage-3} & 
\begin{tabular}[c]{@{}l@{}}
\textbf{Training Data}: 198K \\ 
\textbf{Trainable Module}: Full model
\end{tabular} \\ \hline
\textbf{Training Data (\#Samples)} & 2074K = 1211K + 665K + 198K \\ \hline
\textbf{Training Schedule} & 
\begin{tabular}[c]{@{}l@{}}
\textbf{Learning Rate}: \\
- LLM: 2e-5 \\ 
- Vision Encoder: 2e-6 \\ 
\textbf{Training Epochs}: \\
- Stage-1: 1 epoch \\ 
- Stage-2: 1 epoch \\ 
- Stage-3: 3 epochs \\ 
\textbf{Batch Size}: 128
\end{tabular} \\ \hline
\end{tabular}
\caption{Model Architecture and Training Details}
\label{tab:hyperparameters}
\end{table*}

As illustrated in Table~\ref{tab:hyperparameters} and Figure~\ref{fig:training_stages}, we progressively fine-tune our model with three stages, including language-image alignment and news domain alignment, visual instruction tuning, and misinformation tuning.

To capture detailed visual information for subtle artifact detection, TRUST-VL adopts a dynamic, high-resolution image encoding strategy proven effective in recent VLMs~\cite{li2024llavanext,xue2024xgenmm}. This approach employs patch-wise image encoding, where the original high-resolution image is partitioned into multiple smaller patches, each individually encoded. These patch-level encodings are then concatenated with a downsized version of the original image that provides global contextual information. We utilize the pre-trained CLIP encoder \cite{radford2021learning} to obtain visual representations. 
To align pretrained LLMs with visual encoders, we use lightweight MLP projectors \cite{liu2023visual,liu2024improved} to connect image features into the word embedding space of the language model and then fine-tune the model on instruction-formatted datasets to improve generalization and controllability. 
The language tokens consist of a system message, task-specific instruction, input text, retrieved evidence, and targeted questions.

In our experiments, we use the following model checkpoints as baselines: blip2-flan-t5-xl, instructblip-vicuna-13b, llava-v1.5-13b, llava-v1.6-mistral-13b-hf, xgen-mm-phi3-mini-instruct-r-v1, and Qwen2-VL-7B-Instruct.
For detectors such as MMD-Agent and LRQ-FACT, we utilize llava-v1.5-13b as the VLM for fair comparison.

\section{Datasets}

To evaluate the effectiveness of multimodal misinformation detection models, we leverage a diverse set of in-domain and out-of-domain datasets covering textual, visual, and cross-modal misinformation. These datasets enable a comprehensive assessment of misinformation detection across different modalities and manipulation techniques.

\begin{itemize}[itemsep=.8pt, topsep=2pt, leftmargin=1em]
    \item \textbf{MMFakeBench} \cite{liu2025mmfakebench} is a multimodal misinformation detection benchmark designed to evaluate robustness against various manipulation techniques. It contains 1,000 instances with a distribution of real samples and manipulated cases, including textual veracity distortions, visual veracity distortions, and cross-modal consistency distortions. The dataset introduces 12 forgery types, making it a comprehensive benchmark for evaluating multimodal misinformation detection.

    \item \textbf{Factify2} \cite{suryavardan2023factify2} is a multimodal fact-checking dataset comprising 50,000 instances of supporting and refuting claims sourced from fact-checking platforms such as PolitiFact. This dataset extends the original Factify dataset by incorporating a wider range of real and manipulated news content,  including satirical articles.

    \item \textbf{DGM\textsuperscript{4}-Face} \cite{shao2023dgm} is a large-scale dataset generated by two image manipulation and two text manipulation approaches, with the objective of detecting and grounding manipulations in image-text pairs of human-centric news.
    The original dataset consists of a total of 230K news samples, including 77,426 pristine image-text pairs and 152,574 manipulated pairs. We randomly sample 467 real images and 433 manipulated instances, including face swaps and face attribute modifications.

    \item \textbf{NewsCLIPpings} \cite{luo2021newsclippings} is the largest synthetic benchmark for detecting out-of-context (OOC) misinformation. It generates OOC samples by replacing images in original image-caption pairs with real and semantically related images from different news events. \cite{abdelnabi2022open} further extends this dataset by incorporating textual and visual evidence retrieved via Google Search APIs to improve detection performance.

    \item \textbf{MOCHEG} \cite{yao2023mocheg} is a large-scale dataset for fact-checking, comprising 15,601 claims, each annotated with a truthfulness label and a ruling statement. It includes 33,880 paragraphs and 12,112 images as evidence. It is sourced from fact-checking platforms and serves as a benchmark for evaluating the ability of models to verify textual claims. For fair evaluation, we sample 400 news instances with a balanced distribution of real and fake samples.

    \item \textbf{Fakeddit} \cite{nakamura2020fakeddit} is a large-scale multimodal fake news dataset collected from Reddit. It contains over 1 million instances across multiple categories of misinformation, providing fine-grained 2-way, 3-way, and 6-way classification of fake news. 
    Similarly, we sample 400 news instances with an equal number of real and fake claims.

    \item \textbf{VERITE} \cite{papadopoulos2024verite} is a real-world  dataset designed for detecting out-of-context misinformation, which effectively mitigates the problem of unimodal bias and provides a more robust and reliable evaluation framework. A balanced subset of 400 samples is used to ensure fair evaluation.
\end{itemize}

\lstset{escapeinside=`}

\begin{figure*}[t!]
\lstset{style=mystyle,
        frame=none,
        keywordstyle = \color{black}, 
        commentstyle =\color{codegreen}, 
        stringstyle = \color{black}, 
        breakindent=0\textwidth,
        frame = single,
        backgroundcolor=\color{white},
        xleftmargin=0.05\textwidth,
        xrightmargin=0.05\textwidth}
\begin{lstlisting}[language=Python, numbers=none]
# system message
Task description: some rumormongers intentionally write fake news, manipulate images, or use images from other news events to make multimodal misinformation. Given a news text and a news image, you are responsible for judging whether the given text and image are both credible and faithfully represent the news event. You will be presented with a text and an image. You should use the following step-by-step instructions to derive your judgment: 
# shared steps
Step 1 - Analyze the text: Carefully review the provided text, summarize its key facts, events, and entities. Pay attention to any misleading, false, or fabricated contents. 
Step 2 - Provide a detailed description of the news image: Identify the main subjects, such as people, groups, or specific elements related to the news event. 
# specialized steps
Step 3 -...
# conclusion
Step 6 - What is your final judgment? According to the previous steps, you will first think out loud about your eventual conclusion, enumerating reasons why the news does or does not contain false information. After thinking out loud, you should output either 'Real' or 'Fake' depending on whether you think the given text and accompanying image are both truthful and consistent: 'Real' if the news is factually correct and the image faithfully represent the news text, or 'Fake' if the news is misleading, manipulated or the image is used out of context. 

# input
(*@\color{pink}<image>@*)
Claim Text: (*@\color{pink}<text>@*)  
Direct Evidence: (*@\color{pink}<direct evidence>@*)
Inverse Evidence: (*@\color{pink}<inverse evidence>@*)
Context Evidence: (*@\color{pink}<context evidence>@*)
Your judgment:


\end{lstlisting}
\caption{Prompt used to ask GPT-4o to generate the instruction data.}
\label{fig:gpt4o_prompt}
\end{figure*}

\lstset{escapeinside=`}

\begin{figure*}[t!]
\lstset{style=mystyle,
        frame=none,
        keywordstyle = \color{black}, 
        commentstyle =\color{codegreen}, 
        stringstyle = \color{black}, 
        breakindent=0\textwidth,
        frame = single,
        backgroundcolor=\color{white},
        xleftmargin=0.05\textwidth,
        xrightmargin=0.05\textwidth}
\begin{lstlisting}[language=Python, numbers=none]
# system message
You are a misinformation detection assistant. Task description: some rumormongers intentionally write fake news, manipulate images, or use images from other news events to make multimodal misinformation. Given a news text and a news image, you are responsible for judging whether the given text and image are both credible and faithfully represent the news event. You will be presented with a text, an image, direct evidence, and inverse evidence. For final judgment, you should output either 'Real' or 'Fake' depending on whether you think the given text and accompanying image are both truthful and consistent: 'Real' if the news is factually correct and the image faithfully represent the news text, or 'Fake' if the news is misleading, manipulated or the image is wrongly used in the news text. 

A few rules: 
- If a specific type of evidence (i.e., direct, or inverse) is not provided, state clearly: 'There is no {type} evidence.' 
- Do not nitpick over the direct and inverse evidence as it may contain some noise. 
- Your judgment must always end with either 'Real' or 'Fake'.

# input
(*@\color{pink}<image>@*)
Claim Text: (*@\color{pink}<text>@*)  
Direct Evidence: (*@\color{pink}<direct evidence>@*)
Inverse Evidence: (*@\color{pink}<inverse evidence>@*)
Context Evidence: (*@\color{pink}<context evidence>@*)
Your judgment:


\end{lstlisting}
\caption{TRUST-VL language input.}
\label{fig:trust-vl_prompt}
\end{figure*}

\section{Baselines}
\begin{itemize}[itemsep=.8pt, topsep=2pt, leftmargin=1em]

    \item \textbf{BLIP-2} \cite{li2023blip2} is a vision-language model that bridges the modality gap between vision and language models without requiring training from scratch. It employs a Querying Transformer to effectively align visual features with language models.

    \item \textbf{InstructBLIP} \cite{dai2023instructblip} is an instruction-tuned version of BLIP-2, designed to handle a wide range of vision-language tasks through instruction tuning. By integrating visual instruction tuning, InstructBLIP achieves improved performance across various tasks, including image captioning and visual question answering. 
    
    \item \textbf{LLaVA} \cite{liu2023visual} is one of the pioneering works in visual instruction tuning. It improves the vision-language connector’s representation power with a two-layer MLP to enhance multimodal capabilities.
    
    \item \textbf{LLaVA-NeXT} \cite{li2024llavanext} is an enhanced version of LLaVA
    with improved vision-language alignment and reasoning. It builds upon the original LLaVA framework to offer more accurate and contextually relevant responses in multimodal interactions. 
    
    \item \textbf{xGen-MM} \cite{xue2024xgenmm} also known as BLIP-3, is a large multimodal model framework which replaces the complex Q-Former module used in BLIP-2 with a scalable vision token sampler, specifically a perceiver resampler, to process visual inputs. Additionally, xGen-MM is able to handle free-form interleaved sequences of images and text by adopting a single auto-regressive loss function.
    
    \item \textbf{Qwen2-VL} \cite{wang2024qwen2vl} is a VLM that integrates visual understanding with language processing capabilities. It introduces two key innovations: Naive Dynamic Resolution, allowing the model to process images of varying resolutions by dynamically adjusting the number of visual tokens, and Multimodal Rotary Position Embedding (M-RoPE), which facilitates the effective fusion of positional information across text, images, and videos.

    \item \textbf{GPT-4o} \cite{openai2024gpt4o}. This is currently one of the most powerful multimodal large language models. We utilize GPT-4o in a zero-shot manner with step-by-step instructions for multimodal misinformation detection. 

    \item \textbf{o1} \cite{openai2024o1} is the latest multimodal VLM with advanced reasoning capabilities via large-scale reinforcement learning. For fair comparison, we adopt o1 using the same evaluation protocol as GPT-4o.
    
    \item \textbf{SNIFFER} \cite{qi2024sniffer}.  This is the state-of-the-art large VLM designed for OOC misinformation detection. It employs a two-stage instruction tuning on InstructBLIP \cite{dai2023instructblip} for the cross-modal consistency checks. 
    
    \item \textbf{MMD-Agent} \cite{liu2025mmfakebench} is a multimodal agent framework that integrates the reasoning, action, and tool-use capabilities of LVLM agents. It decomposes misinformation detection into three sequential stages: textual veracity check, visual veracity check, and cross-modal consistency reasoning. This structured approach enables systematic and thorough analysis. At each stage, MMD-Agent prompts LVLMs to generate multi-perspective reasoning traces and coordinates their outputs to obtain a final decision.
    
    \item \textbf{LRQ-FACT} \cite{beigi2024lrqfact} is a fact-checking system that utilizes a multi-agent framework to leverage VLMs and LLMs to generate comprehensive questions and answers for understanding multimodal content. Then, a decision-maker LLM assesses the veracity based on all generated context.

\end{itemize}

\section{Model Prompts}

Our structured reasoning template is designed to reflect widely adopted human fact-checking workflows, which typically involve decomposed, step-by-step verification processes \cite{nakov2021automated,vlachos-riedel-2014-fact,warren2025show}. Prior studies have formalized fact-checking as a pipeline involving claim analysis, evidence retrieval, consistency assessment, and final verdict prediction. For example, \cite{warren2025show} highlights that professional fact-checkers require transparent, explainable systems that mirror their multi-stage decision-making processes.

\begin{table*}[tp!]
\small
\centering
\begin{tabular}{lcccc}
\toprule
Model & MMFakeBench & Factify2 & DGM$^{4}$-Face & NewsCLIPpings \\
\midrule
TRUST-VL-7B (Backbone: LLaVA)   & 85.90 & 99.33 & 80.90 & 88.79 \\
TRUST-VL-7B (Backbone: Mistral) & 85.70 & 99.30 & 82.12 & 88.53 \\
\bottomrule
\end{tabular}
\caption{Performance (\%) of TRUST-VL with different backbone models.
}
\label{tab:backbone_ablation}
\end{table*}

\begin{table}[t!]
\small
\centering
\resizebox{\columnwidth}{!}{
\begin{tabular}{lccccc}
\toprule
Proportion & 0\% & 25\% & 50\% & 75\% & 100\% \\
\midrule
Acc. & 90.35 & 89.09 & 88.54 & 84.10 & 81.96 \\
\bottomrule
\end{tabular}
}
\caption{TRUST-VL’s performance (\%) across varying proportion of incorrect evidence on NewsCLIPpings.}
\label{tab:evidence_robustness}
\end{table}

\begin{table}[t!]
\small
\centering
\resizebox{\columnwidth}{!}{
\begin{tabular}{lcc}
\toprule
Dataset & MMD-Agent (LLaVA) & MMD-Agent (GPT-4o) \\
\midrule
MMFakeBench   & 69.10 & 76.56 \\
Factify2      & 71.03 & 84.00 \\
DGM$^{4}$-Face & 48.30 & 55.96 \\
NewsCLIPpings & 53.06 & 77.34 \\
\bottomrule
\end{tabular}
}
\caption{Performance (\%) comparison of MMD-Agent with different backbones.}
\label{tab:mmd_agent_variants}
\end{table}

Figure \ref{fig:gpt4o_prompt} illustrates the prompt utilized for asking GPT-4o to generate instruction data. For each claim, we retrieve textual and visual evidence (converted to text via image captioning) separately and then pass them to GPT-4o to process. We also consider context evidence provided by users or downstream tasks. For specialized steps, we carefully design critical steps required for addressing different distortion types. Finally, GPT-4o outputs a final judgment along with detailed explanations, guided by carefully designed step-by-step reasoning instructions. Figure \ref{fig:trust-vl_prompt} shows language input for the TRUST-VL framework. Together, these prompt designs ensure high-quality reasoning supervision during training and robust, explainable predictions.

Although the input formats and reasoning templates vary across tasks, our proposed unified model can handle them all. The reasoning template is carefully designed to reflect the inherent characteristics of each distortion type. For instance, tasks involving visual distortion primarily require the model to detect fine-grained visual artifacts in the image modality, where evidence-based reasoning paths are not beneficial to the final judgment. Our unified framework reformulates all tasks into a consistent structure comprising a chain of question-answering steps followed by a final veracity judgment that integrates multiple reasoning paths.

\section{Additional Experiments}

\noindent\textbf{Impact of Backbone Choice.} To demonstrate the generalizability of our proposed framework and instruction data, we further evaluate TRUST-VL using an alternative backbone (Mistral-7B \cite{jiang2023mistral}), as shown in Table~\ref{tab:backbone_ablation}. The results demonstrate that TRUST-VL achieves highly consistent performance across datasets, with comparable accuracy under both LLaVA and Mistral backbones. These findings confirm that the improvements are not tied to any specific backbone.

\noindent\textbf{Effect of Incorrect Evidence.} To examine whether TRUST-VL can still make correct inferences when provided with misleading or incorrect evidence, we randomly sample irrelevant evidence into the input and systematically evaluate the robustness of our proposed model under varying proportions of incorrect evidence (0, 25\%, 50\%, 75\%, 100\%) on NewsCLIPpings, as shown in Table~\ref{tab:evidence_robustness}. Notably, even under a large amount of incorrect evidence (75\%), TRUST-VL maintains strong performance and makes reliable predictions despite noisy evidence (e.g., a 6.25-point drop).

\begin{figure}[t!]
  \centering
  \includegraphics[width=0.85\linewidth]{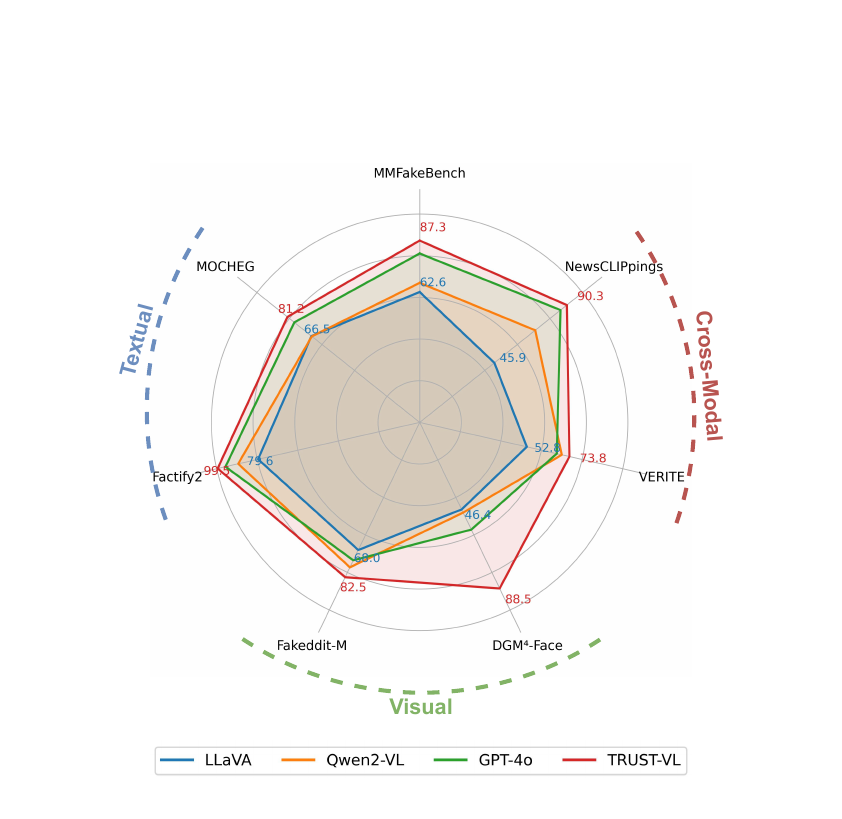}
  \caption{Performance (\%) comparison between TRUST-VL and general VLMs.}
  \label{fig:radar_performance}
\end{figure}

\begin{figure*}[t!]
  \centering
  \includegraphics[width=\linewidth]{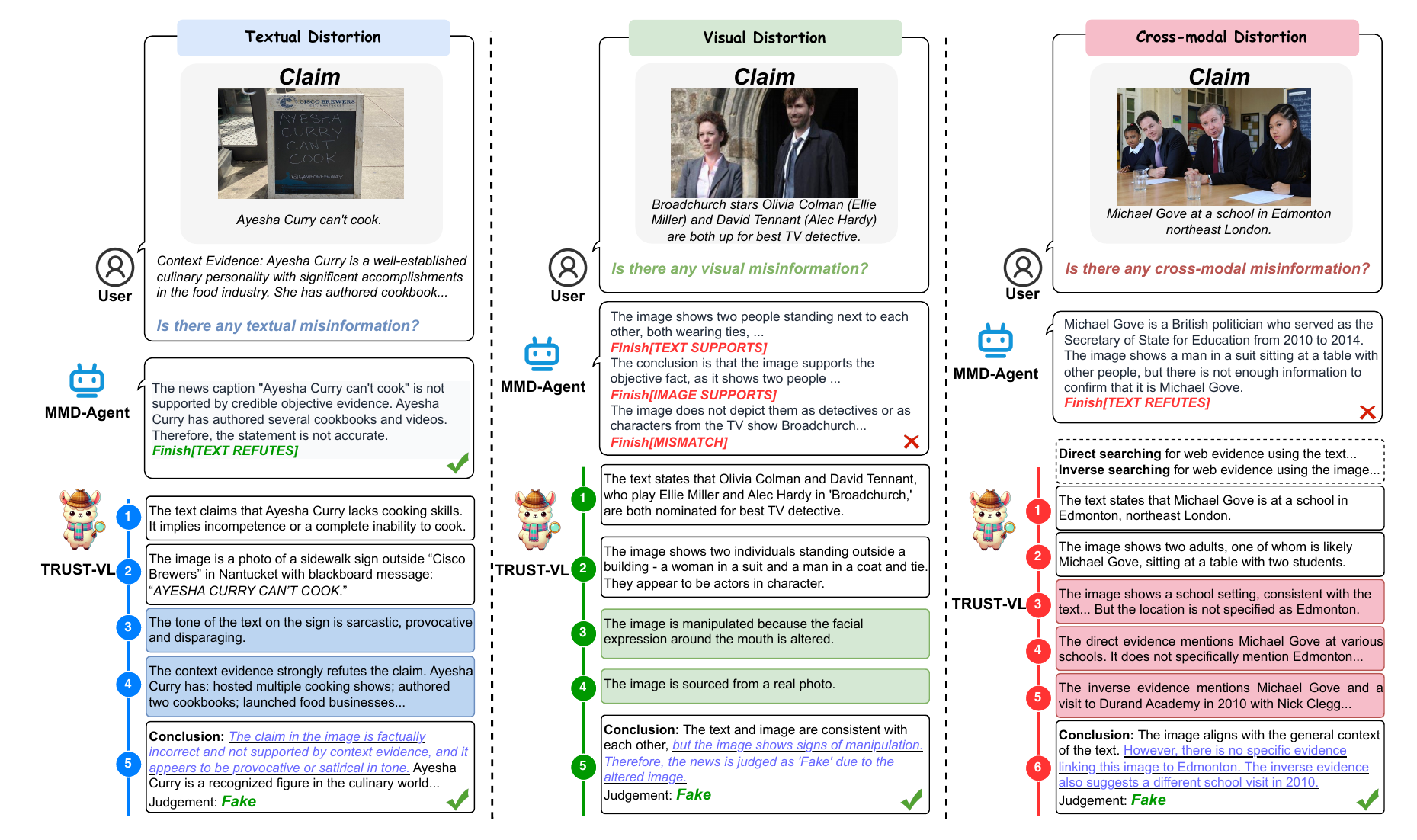}
  \caption{Comparison between the proposed TRUST-VL and specialized detectors.}
  \label{fig:case_3}
\end{figure*}

\noindent\textbf{MMD-Agent Variants} We used llava-v1.5-13b as the vision-language model backbone for MMD-Agent to ensure a fair comparison among open-source baselines. As shown in Table~\ref{tab:mmd_agent_variants}, using GPT-4o as the base backbone significantly improves MMD-Agent’s performance but still performs substantially worse than the proposed TRUST-VL. This discrepancy reveals the sensitivity of MMD-Agent to the capabilities of its base models. As illustrated in Figure~\ref{fig:radar_performance}, existing vision-language models, including GPT-4o, struggle with subtle visual manipulations, particularly in tasks like DGM\textsuperscript{4}-Face. Additionally, we observed that MMD-Agent frequently suffers from incorrect grounding in its sequential reasoning process. This often leads to an early stop and incomplete verification, which degrade its detection performance.

\section{Additional Case Study}

Figure~\ref{fig:case_3} showcases three real-world misinformation cases, each demonstrating a distinct distortion type: textual, visual, and cross-modal. Specialized misinformation detectors such as MMD-Agent tend to produce shallow or incomplete assessments. For instance, in the Ayesha Curry case, it offers a brief factual correction without recognizing the satirical tone; in the Olivia Colman case, it fails to detect the subtle visual manipulation; and in the third case, it misidentifies the setting despite contradictory evidence. These limitations highlight MMD-Agent’s lack of in-depth reasoning and explainability, especially when dealing with subtle visual manipulations or cross-modal distortions, which TRUST-VL addresses more effectively.

\end{document}